\documentclass[lettersize,journal]{IEEEtran}
\usepackage{amsmath,amsfonts}
\usepackage{algorithmic}
\usepackage{algorithm}
\usepackage{array}
\usepackage[caption=false,font=normalsize,labelfont=sf,textfont=sf]{subfig}
\usepackage{textcomp}
\usepackage{stfloats}
\usepackage{url}
\usepackage{verbatim}
\usepackage{graphicx}
\usepackage{cite}

\usepackage{multirow}
\usepackage{latexsym}
\usepackage{amssymb}
\usepackage{amsthm}
\usepackage{microtype}
\usepackage{amsmath}
\usepackage{makecell}
\usepackage{xcolor}
\usepackage{soul}
\usepackage{bbding}
\usepackage{caption}

\definecolor{Seashell}{RGB}{255,228,181} 
\definecolor{Firebrick4}{RGB}{210,105,30}
\newtheorem*{exmp}{Example}

\definecolor{Redshellshallow}{RGB}{253,245,230} 
\definecolor{Redbrick4}{RGB}{0,0,0}
\newcommand{\coderedshallow}[1]{
  \begingroup
  \sethlcolor{Redshellshallow}
  \textcolor{Redbrick4}{\hl{#1}}
  \endgroup
}

\definecolor{Redshellmiddle}{RGB}{255,222,173} 
\definecolor{Redbrick4}{RGB}{0,0,0}
\newcommand{\coderedmiddle}[1]{
  \begingroup
  \sethlcolor{Redshellmiddle}
  \textcolor{Redbrick4}{\hl{#1}}
  \endgroup
}

\definecolor{Redshellhigh}{RGB}{255,165,0} 
\definecolor{Redbrick4}{RGB}{0,0,0}
\newcommand{\coderedhigh}[1]{
  \begingroup
  \sethlcolor{Redshellhigh}
  \textcolor{Redbrick4}{\hl{#1}}
  \endgroup
}

\definecolor{Blueshell}{RGB}{100,149,237} 
\definecolor{Bluerick4}{RGB}{0,0,128}

\begin{document}

\title{Emotion Prediction Oriented method with Multiple Supervisions for Emotion-Cause Pair Extraction}


\author{Guimin Hu, Yi Zhao, Guangming Lu
\thanks{
This work was supported by the Innovative Research Project of Shenzhen under Grant No. KQJSCX20180328165509766 and the Natural Science Foundation of Guangdong, China under Grant No. 2020A1515010812 and 2021A1515011594.(Corresponding author:Yi Zhao).\\
Guimin Hu and Guangming Lu are with the School of Computer Science, and Technology, Harbin Institute of Technology (Shenzhen), Shenzhen 518055, China, (e-mail: rice.hu.x@gmail.com; luguangm@hit.edu.cn).}
\thanks{Yi Zhao is with the School of Science, Harbin Institute of Technology (Shenzhen), Shenzhen 518055, China, (e-mail: zhao.yi@hit.edu.cn)}
}

\markboth{Journal of \LaTeX\ Class Files,~Vol.~14, No.~8, August~2021}%
{Shell \MakeLowercase{\textit{et al.}}: Emotion Prediction Oriented method with Multiple Supervisions for Emotion-Cause Pair Extraction}


\maketitle

\begin{abstract}
Emotion-cause pair extraction (ECPE) task aims to extract all the pairs of emotions and their causes from an unannotated emotion text. The previous works usually extract the emotion-cause pairs from two perspectives of emotion and cause. However, emotion extraction is more crucial to the ECPE task than cause extraction. Motivated by this analysis, we propose an end-to-end emotion-cause extraction approach oriented toward emotion prediction (EPO-ECPE), aiming to fully exploit the potential of emotion prediction to enhance emotion-cause pair extraction. Considering the strong dependence between emotion prediction and emotion-cause pair extraction, we propose a synchronization mechanism to share their improvement in the training process. That is, the improvement of emotion prediction can facilitate the emotion-cause pair extraction, and then the results of emotion-cause pair extraction can also be used to improve the accuracy of emotion prediction simultaneously. For the emotion-cause pair extraction, we divide it into genuine pair supervision and fake pair supervision, where the genuine pair supervision learns from the pairs with more possibility to be emotion-cause pairs. In contrast, fake pair supervision learns from other pairs. In this way, the emotion-cause pairs can be extracted directly from the genuine pair, thereby reducing the difficulty of extraction. Experimental results show that our approach outperforms the 13 compared systems and achieves new state-of-the-art performance.
\end{abstract}


\section{Introduction}
The purpose of sentiment analysis \cite{DBLP:conf/naacl/WilsonWH05,DBLP:journals/ftir/PangL07,DBLP:conf/acl/LiHWZ15,DBLP:conf/acl/QianHLZ17,DBLP:conf/emnlp/HuLZLWL22} is to classify a given text into positive, negative, neutral or more fine-grained categories, which usually plays an important role in decision-making and human behavior analysis. Recently, more and more studies \cite{DBLP:conf/nlpcc/HuLZ20,DBLP:conf/emnlp/ChenLW20,DBLP:journals/taslp/FanYGZX21} have focused not only on the emotional polarity of a text but also on the reasons behind this emotion. This promotes the proposal of the emotion-cause extraction task (ECE) and the emotion-cause pair extraction task (ECPE). Emotion cause analysis has attracted lots of attention due to the promising application prospects. Generally, the emotion clause appears in text with an emotional word, such as ``happiness'' and ``sadness'', and the cause clause is expressed as an event or a kind of behavior. The cause provides more details about how emotions are stimulated and generated. Extracting emotion causes help to understand what emotions are expressed in the text and why these emotions are perceived. The emotion cause analysis can be applied in many fields, such as improving service quality \cite{DBLP:conf/acl/OhLWPSKK20,DBLP:conf/acl/YanDJQ020,DBLP:conf/emnlp/XuLLB20}, controlling public opinion trends \cite{DBLP:journals/corr/abs-2010-04640,DBLP:conf/emnlp/DingXY20,DBLP:journals/corr/abs-2002-10710,DBLP:conf/naacl/KeXL21} and assisting the empathetic response generation \cite{DBLP:conf/emnlp/GaoLDWCDX21,DBLP:conf/emnlp/Kim0K21}. Particularly, emotion cause extraction (ECE) aims to determine which clauses contain the causes for the given emotion \cite{lee-etal-2010-text,DBLP:conf/emnlp/GuiWXLZ16,DBLP:conf/emnlp/LiSFWZ18}. \cite{DBLP:conf/acl/XiaD19} points out that the ECE task neglects the mutual indications between the emotion and cause and requires the annotated emotion in advance, which limits its application scope. To solve these shortcomings, \cite{DBLP:conf/acl/XiaD19} define a new emotion-cause pair extraction (ECPE) task, which aims to extract all the potential pairs of emotions and the corresponding causes from an unannotated emotion text. As shown in the following example, an emotion clause $c_{2}$ containing the emotion word ``happily'' and the corresponding cause clause $c_{1}$ constitute an emotion-cause pair $(c_{2},c_{1})$:
\begin{exmp}
He received a call to be hired ($c_{1}$), jumped up happily ($c_{2}$), and told his parents this good news ($c_{3}$).
\end{exmp}

For the ECPE task, \cite{DBLP:conf/acl/XiaD19} firstly proposed a two-step method, where the first step aims to extract emotions and cause clauses separately. The second step is to train a binary classifier to determine whether a candidate pair is an emotion-cause pair. Obviously, this two-step setting will cause the errors in the first step to propagate to the second step. To address this problem, some works proposed end-to-end architectures to solve the ECPE task. Some works \cite{DBLP:conf/acl/DingXY20,DBLP:journals/corr/abs-2002-10710,DBLP:conf/acl/WeiZM20} learned the representation of candidate pair and made classification on the pair representation to identify the emotion-cause pairs in a unified fashion. Some works \cite{DBLP:conf/emnlp/DingXY20,DBLP:conf/coling/ChengJYYG20} constructed the networks from the perspectives of emotion and cause and extracted the corresponding cause clause/emotion clause based on the assumption that each clause in the document can be regarded as an emotion clause/cause clause. Furthermore, some works \cite{DBLP:conf/emnlp/YuanFBX20,DBLP:conf/coling/ChenLW20,DBLP:journals/taslp/ChengJYLG21,DBLP:journals/taslp/FanYGZX21} regarded the ECPE as a sequence labeling problem. \cite{DBLP:conf/emnlp/YuanFBX20} considered the relative distance between the emotion clause and cause clause into tags. \cite{DBLP:conf/coling/ChenLW20} added the emotional type to labeling tag. \cite{DBLP:journals/taslp/ChengJYLG21} designed content and pairing parts for emotion/cause identification and clause pairing, respectively, and propose a unified target-oriented sequence-to-sequence model to capture the mutual effects among the target clause, global context, and former decoded label. \cite{DBLP:journals/taslp/FanYGZX21} proposed a multi-task sequence tagging framework and used the output of both auxiliary tasks to induce the tag distribution for benefiting emotion-cause pair extraction. 

Determining the emotion clause is essential for emotion-cause pair extraction \cite{DBLP:conf/acl/XiaD19,DBLP:conf/coling/ChengJYYG20,DBLP:conf/emnlp/DingXY20}. However, how to effectively construct and utilize the high-quality emotion clause candidates to improve the performance of emotion-cause pair extraction still lack sufficient exploration. Actually, when solving ECPE, the emotion clauses are usually easier to be detected in comparison with the cause clauses. Suppose we can preliminarily determine the possible emotion clauses and find the probable cause clauses for each likely emotion clause. In that case, we can obtain a candidate set containing the pairs with a high possibility of becoming emotion-cause pairs. The extracted emotion clause can provide clues to assist cause clause extraction, which reduces extraction difficulties as we can extract real emotion-cause pairs from these candidate pairs with a higher chance. Motivated by this idea, we propose an end-to-end framework oriented toward emotion prediction (EPO-ECPE) to extract the emotion-cause pairs, which centers around the emotion prediction to guide the subsequent emotion-cause pair extraction. Obviously, the performance of emotion-cause pair extraction is impacted by the quality of the candidate emotion clause. That is, the low precision and low recall of emotion prediction will damage the performance of emotion-cause pair extraction, and the high performance of emotion prediction is conducive to constructing a higher-quality emotion-cause pair candidate set. We equip EPO-ECPE with a synchronization mechanism to exploit the potential of emotion prediction in assisting emotion-cause pair extraction. This ensures that the knowledge learned from the emotion prediction and emotion-cause pair extraction can be synchronized to promote each other. Different candidate pairs have different probabilities of being emotion-cause pairs. Based on this observation, we divide all pairs into pairs that are more likely to be emotion-cause pairs, i.e., genuine pairs, and pairs that are less likely to be emotion-cause pairs, i.e., fake pairs.
Meanwhile, we set the genuine and fake pair supervisions to learn from the genuine and fake pairs, respectively. The two supervisions update the pair representations in the pairing stage in different ways and then share the learned representations with the emotion prediction to improve a new round of emotion prediction. Note that the new round of emotion prediction can further assist the emotion-cause pair extraction.  
The main contributions of this paper are summarized as follows:
\begin{itemize}
\item[1.] We propose an end-to-end emotion-cause extraction approach oriented toward emotion prediction (EPO-ECPE). To our knowledge, it is the first ECPE framework centered on emotion prediction to guide subsequent emotion-cause pair extraction.

\item[2.] We propose a synchronization mechanism to share the learned clause representation between emotion prediction and emotion cause pair extraction in multiple training rounds. To reduce the difficulty of pair extraction, we divide all candidate pairs into genuine pairs and fake pairs and set the genuine pair supervision and fake pair supervision to gather information from the genuine pairs and fake pairs, respectively.

\item[3.] Experimental results demonstrate that the proposed EPO-ECPE outperforms all existing methods and achieves state-of-the-art performance on emotion-cause pair extraction. Especially, EPO-ECPE greatly improves when extracting multiple emotion-cause pairs from one document \footnote{https://github.com/LeMei/EPO-ECPE}.
\end{itemize}

\section{Related Work}

\subsection{Emotion Cause Extraction}

\cite{lee-etal-2010-text} gave the initial definition of emotion cause extraction task (ECE) to extract the word-level causes of the given emotion in text. Early works of ECE mostly were dependent on linguistic rules \cite{lee-etal-2010-text,DBLP:conf/coling/ChenLLH10,DBLP:conf/nlpcc/GuiYXLLZ14,DBLP:conf/pakdd/GaoXW15} or traditional machine learning algorithms \cite{DBLP:conf/emnlp/GuiWXLZ16}. With the development of the deep neural network, the recent works of ECE task gradually adopted the deep learning models, such as co-attention \cite{DBLP:conf/emnlp/LiSFWZ18}, self-attention \cite{DBLP:conf/ijcai/XiaZD19}, multi-task learning \cite{DBLP:conf/emnlp/ChenHCL18}, graph neural networks \cite{DBLP:journals/kbs/HuLZ21}, and external knowledge incorporation \cite{DBLP:conf/acl/Yan0PH20}, to solve this task. Specifically, \cite{DBLP:journals/corr/abs-1708-05482} proposed a convolutional multiple-slot memory network that simultaneously captures each word's context. \cite{DBLP:conf/emnlp/ChenHCL18} unified the emotion classification and cause detection to fully exploit the interaction between emotion classification and cause detection. \cite{DBLP:conf/emnlp/LiSFWZ18} considered the emotional context awareness to construct the co-attention network to incorporate the emotional information into the clause representation. \cite{DBLP:conf/aaai/DingHZX19} viewed ECE as a reordered prediction problem and took previous clauses' prediction labels into the subsequent clauses' subsequent prediction. \cite{DBLP:conf/ijcai/XiaZD19} adopted Transformer \cite{DBLP:conf/nips/VaswaniSPUJGKP17} as the clause encoder to encode the mutual indication among clauses in a document to extract the emotion cause clause based on such global information. \cite{DBLP:journals/kbs/HuLZ21} constructed a graph structure based on the inter-clause dependency and proposed a graph convolutional network with the fusion of semantics and structural information (FSS-GCNs), which captures the semantic relevance and structural constricts among clauses simultaneously. \cite{DBLP:journals/kbs/XuLLX21} proposed a two-stage supervised method from the perspective of information retrieval ranking to extract the emotion causes. It distinguishes the causes for each emotion in contexts by using the query performance predictors in the first stage and then considers the emotional complexity to enhance the ranking model for accurately extracting the causal clauses in the second stage. \cite{DBLP:conf/emnlp/FanYDGBYXM19} used the hierarchical RNN to model the document structure and injected the sentiment lexicon and common knowledge into the model in a regularized way to constrain the model parameters. \cite{DBLP:conf/emnlp/HuLZ21} utilized the hierarchy and bidirectionally of context to focus on the relevant contextual information to the candidate cause clause and incorporate that information as the features for detecting the emotion causes. \cite{DBLP:conf/ijcnn/LinYG21} proposed a Hierarchical Inter-Clause Interaction Network (HICIN) network to capture the inter-clause interaction on both word-level and clause-level, which captures the semantic cues at multiple granularities. \cite{DBLP:journals/kbs/HuZLYC22} formalized ECPE as a probability problem and evaluated the mutual information between emotion clause and cause clause. Furthermore, \cite{DBLP:journals/kbs/HuZLYC22} proved the conjecture on the emotion-cause causality and mutual information.
\cite{DBLP:conf/acl/Yan0PH20} utilized commonsense knowledge to construct the emotion trigger path between candidate clauses and emotion clauses and proposed a novel strategy to generate adversarial examples to alleviate the position bias of the benchmark dataset.

\subsection{Emotion-Cause Pair Extraction}
Although the ECE task has achieved significant progress, it suffers from two shortcomings: 1) the emotion must be annotated before extracting the cause clause in the ECE task, and 2) the fact that the emotion and its corresponding cause clause are mutually indicative are ignored. The first dramatically limits its applications in real-world scenarios, and the second lacks the utilization of mutual indication between emotion and cause.
To address these shortcomings, more recently, emotion-cause pair extraction (ECPE) has developed based on the ECE task. ECPE task aims to extract the potential emotions and the corresponding causes together from an unannotated text \cite{DBLP:conf/acl/XiaD19} in the form of clause pairs. For the ECPE, \cite{DBLP:conf/acl/XiaD19} proposed a two-step architecture, where the first step is to construct the interactive network between the emotion extraction and cause extraction to extract the emotions and causes simultaneously. The second step is to filter the negative pair by classifying the pairs constructed by emotion and cause. This two-step method is a pipeline system, which may lead to the further propagation of the error from the first step to the second step. To solve this problem, \cite{DBLP:conf/acl/WeiZM20} adopted a unified architecture to encode the pair representation by modeling the inter-clause dependency and determined emotion-cause pairs from the ranking perspective. \cite{DBLP:conf/acl/DingXY20} proposed 2D-Transformer to integrate the representation, interaction, and prediction into a joint framework to solve the ECPE. \cite{DBLP:conf/emnlp/DingXY20} assumed that all clauses can be viewed as the emotion clauses and cause clauses separately and then extracted their causes and emotions. \cite{DBLP:conf/acl/FanYDGYX20} viewed the ECPE as a parsing-like directed graph construction procedure and generated the directed graph with labeled edges based on a sequence of actions. Some works \cite{DBLP:conf/emnlp/YuanFBX20,DBLP:conf/coling/ChenLW20} used the sequence labeling methods to label emotion-cause pairs. \cite{DBLP:journals/taslp/ChengJYLG21} solved the emotion-cause pair extraction task from the perspective of the sequence labeling and designed a content tagging to identify the emotion/cause and a pairing tag to pair the clauses, which utilizes the information of target clause, global context, and former decoded label, to form clause representation. \cite{DBLP:journals/taslp/FanYGZX21} proposed a multi-task sequence tagging framework and used the prediction distribution of both auxiliary tasks as an inductive bias to refine the tag distribution. \cite{DBLP:conf/emnlp/ChenLW20} propose a new task to determine whether the emotion and cause clauses have valid causality in different contexts and construct a corresponding dataset via manual annotation and negative sampling based on an existing benchmark dataset.

Unlike the previous works, we observed that emotion clause extraction is relatively more accessible than cause clause extraction. Still, it is more critical for subsequent emotion-cause pair extraction. Based on this observation, we propose a unified framework oriented toward emotion prediction (EPO-ECPE) and establish a synchronization mechanism between the emotion prediction and subsequent pairing to fully use each module's benefits.

\begin{figure*}[!t]
\centerline{\includegraphics[scale=0.45]{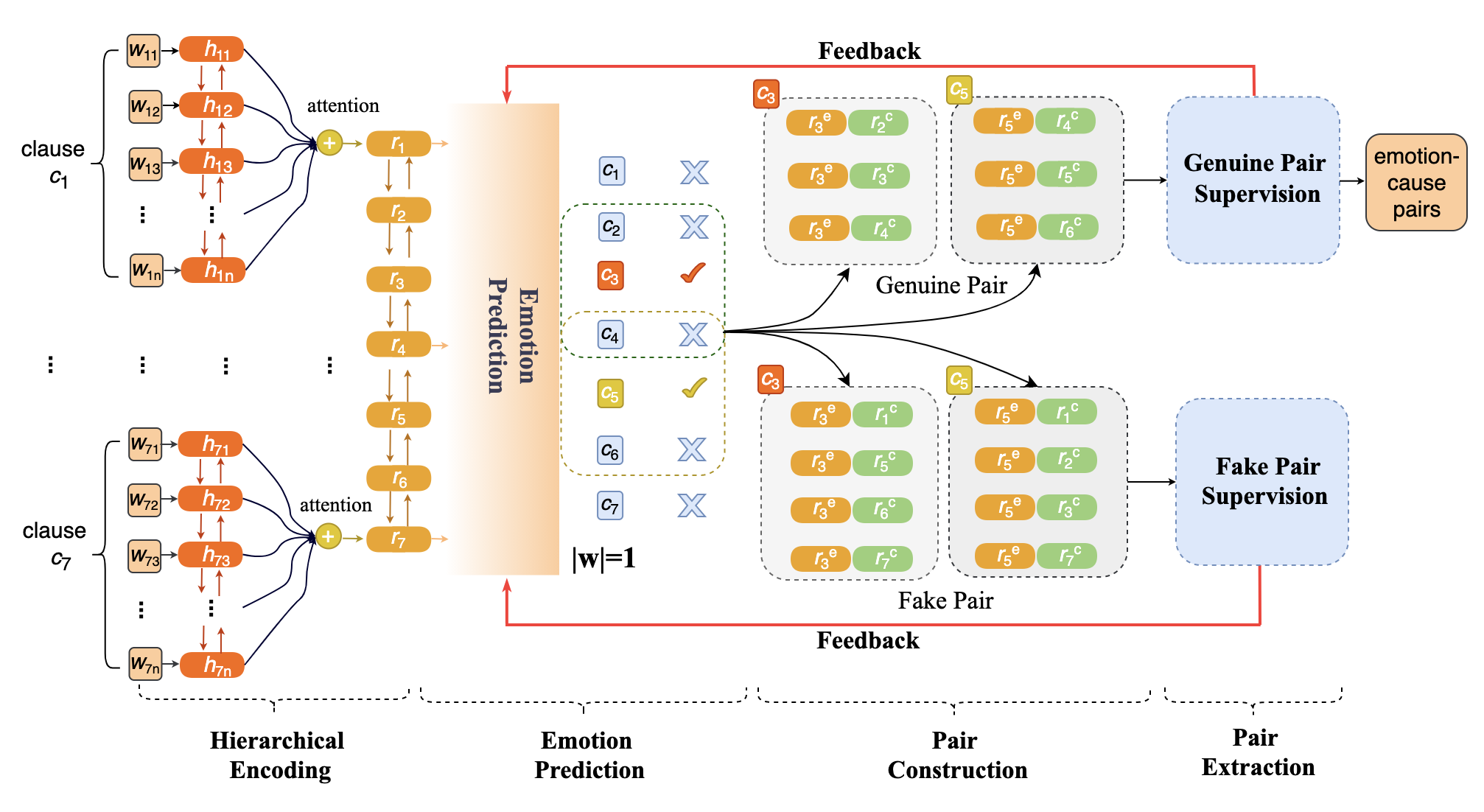}}
\caption{Overview of the EPO-ECPE model shown with a document $d=\{c_{1},...,c_{7}\}$. The hierarchical encoder is used to capture the sequence relations at the word and clause levels to form the clause representations.
After the hierarchical encoder, the emotion prediction preliminarily determines $c_{3}$ and $c_{5}$ are the candidate emotion clauses. Then the genuine pair supervision and fake pair supervision learn from the genuine pairs and fake pairs constructed based on candidate emotion clauses, respectively. Both supervisions update the clause representations by the synchronization mechanism to improve the new emotion prediction round. Here, we set $|w|=1$ and $K=2$ for simplification.}
\label{fig:architecture}
\end{figure*}
\section{Approach}

\subsection{Motivation}
For the benchmark dataset \cite{DBLP:conf/acl/XiaD19}, the average size of document length is 14.77, and each document contains no more than 4 emotion-cause pairs. Given a document containing $|d|$ clauses, the previous works directly extract emotion-cause pairs (4 emotion-cause pairs at most) from the $(|d|*|d|)$ candidate pairs constructed by the Cartesian product of clauses. Extracting a few emotion-cause pairs from many candidate pairs is challenging. But suppose we preliminarily determine which clauses are more likely to be emotion clauses in advance, and find the possible cause clauses for each candidate emotion clause. In that case, we can roughly obtain those pairs with a high possibility to be emotion-cause pairs, which means that we can find the real emotion-cause pairs from these candidate pairs with a higher chance. Compared with extracting emotion-cause pairs from $(|d|*|d|)$ pairs, this will significantly reduce the difficulty of ECPE. However, if the previously determined emotion clause candidates are wrong, the subsequent emotion-cause pair extraction becomes meaningless. Taking this drawback into account, we specially equip the EPO-ECPE with a synchronization mechanism to transfer information between emotion prediction and emotion-cause pair extraction. Based on the results of emotion prediction, we divide all candidate pairs into the pairs that are more likely to be emotion-cause pairs, i.e., genuine pairs, and the pairs that are less likely to be emotion-cause pairs, i.e., fake pairs. To fully exploit the feedback of genuine and fake pairs, we employed genuine pair supervision and fake pair supervision to learn from the genuine pairs and fake pairs, respectively. The representations can be synchronized among emotion prediction, genuine pair supervision, and fake pair supervision through the supervision signals.

\subsection{Task Definition}
Given a document $d=\{c_{1},c_{2},...,c_{|d|}\}$ which consists of $|d|$ clauses, the $i$-th clause $c_{i}=\{w_{i1},w_{i2},...,w_{in}\}$ contains $n$ words. Each document has one or more emotions, and each emotion corresponds to at least one cause. Our goal is to extract all the emotion-cause pairs like \{$\cdots$, $(c_{emo}, c_{cau})^{j}$, $\cdots$\} from document $d$, where $(c_{emo}, c_{cau})^{j}$ is the $j$-th emotion-cause pair of $d$ and $c_{emo}$ is an emotion clause containing a certain emotion, such as ``sadness''. These two clauses construct an emotion-cause pair $(c_{emo}, c_{cau})^{j}$.

\subsection{Overview Architecture}
The proposed EPO-ECPE can be decomposed into a hierarchical encoder (i.e., word-level encoder and clause-level encoder), emotion prediction, pair construction, and pair extraction (i.e., genuine pair supervision and fake pair supervision). Figure \ref{fig:architecture} gives the overall architecture of EPO-ECPE. The hierarchical encoder is used to learn clause representation based on the discourse structure of the document. The emotion prediction aims to form an emotion clause candidate set. Given the candidate emotion clauses, we construct the genuine pairs and fake pairs for each candidate emotion clause and set the genuine pair supervision and fake pair supervision to regularize the representation learning of the genuine pairs and fake pairs, respectively.

\subsection{Hierarchical Encoder}
Document-level context exhibit textual structure, which serves as helpful information for clause representation learning. A general method uses a hierarchical RNN encoder to model this structure. Our hierarchical RNN encoder consists of several parts: a word-level encoder, a word-level attention layer, and a clause-level encoder. 

\subsubsection{Word-level Encoder}
For each clause $c_{i}=\{w_{i1},w_{i2},...,w_{in}\}$, we use the word-level Bi-LSTMs (Bidirectional Long Short-Term Memory) \cite{DBLP:journals/neco/HochreiterS97,DBLP:journals/nn/GravesS05} to capture the sequence features and obtain the hidden state sequence $\{h_{i1},h_{i2},...,h_{in}\}$, where $h_{ij}$ is the hidden state of word $w_{ij}$. Each hidden state $h_{ij}$ is the concatenation of the forward hidden state $\overrightarrow{h_{ij}}$ and backward hidden state $\overleftarrow{{h}_{ij}}$:
\begin{equation}
h_{ij} = [\overrightarrow{h_{ij}},\overleftarrow{h_{ij}}], \label{eq_lstm}
\end{equation}
where $\left [ \cdot,\cdot\right ]$ is the concatenation operation.
We apply an attention model \cite{DBLP:journals/corr/BahdanauCB14,DBLP:conf/naacl/YangYDHSH16} to calculate the attention weights and assign more considerable importance to the representations of those words that are important to the clause. Then the representations of those informative words are aggregated to form the clause representation $h_i$ by the following equations:
\begin{equation}
\begin{split}
\alpha_{ij}=\text{Softmax}(\mathrm{tanh}(W^{w}h_{ij}+b^{w})^\mathsf{T}u^{u})\\
{h}_{i}=\sum_{j=1}^{n}\alpha_{ij}{h}_{ij} \quad\quad\quad\quad\quad\quad \label{eq:word}
\end{split}
\end{equation}
where Softmax is the normalization function, and $\{W^{w},b^{w},{u}^{u}\}$ are learnable parameters. $\alpha_{ij}$ is the attention weight after normalization, showing the importance of word $w_{ij}$ to clause $c_{i}$. The $c_{i}$'s hidden state $h_{i}$ is finally obtained based on the attention weights.
\subsubsection{Clause-level Encoder}
The semantic information of a clause is closely related to its neighbor clauses. Based on this consideration, the clause state sequence $\{{h}_{1},{h}_{2},...,{h}_{|d|}\}$ is fed into a clause-level Bi-LSTM to model the latent semantic relations among clauses in a document, and generate the clause representations $\{{r}_{1},{r}_{2},...,{r}_{|d|}\}$.

\subsection{Emotion Prediction}
For the ECPE task, constructing a high-quality emotion clause candidate set is crucial for generating the subsequent emotion-cause pair candidate set. 
In this section, we aim to extract the first $K$ clauses most likely to be emotion clauses as the emotion clause candidate set. We generate an emotion-specific representation ${r}^{e}_{i}$ and a context-specific representation ${r}^{c}_{i}$ respectively by using two linear functions to transform ${r}_{i}$:
\begin{equation}
\begin{split}
{r}^{e}_{i} = {W}^{e}{r}_{i} + {b}^{e}\\
{r}^{c}_{i} = {W}^{c}{r}_{i} + {b}^{c}
\end{split}
\end{equation}
where ${W}^{e}$, ${W}^{c}$, ${b}^{e}$, ${b}^{c}$ are the learnable parameters. 

Based on ${r}^{e}_{i}$, the prediction probability of clause $c_{i}$ being an emotion clause (denoted as $\hat{y}_{i}^{e}$) is given, and the emotion clause candidate set $CE$ is then constructed on this basis:
\begin{equation}
\begin{split}
\hat{y}_{i}^{e} = \sigma({W}^{e}{r}^{e}_{i} + {b}^{e})\quad\quad
\\CE = \mathop{MAX}\limits_{K}(\hat{y}_{1}^{e},\hat{y}_{2}^{e},...,\hat{y}_{|d|}^{e}) \label{eq_ce}
\end{split}
\end{equation}
where $\sigma(\cdot)$ is the logistic function and $\{{W}^{e},{b}^{e}\}$ are the learnable parameters. $\mathop{MAX}\limits_{K}$ is a function to output the clauses with the former $K$ maximum probability.
\subsection{Pair Construction}
We take each emotion clause candidate $c_{i}\in CE$ as the centre to construct $IW_{i}=\{c_{i-|w|},...,c_{i},...,c_{i+|w|}\}$, and denote $IW_{i}$ as the context clauses of $c_{i}$, where $|w|$ is hyperparameter. Meanwhile, we construct $OW_{i}=\{c_{1},...,c_{i-|w|-1},c_{i+|w|+1},...,c_{|d|}\}$ and denote $OW_{i}$ as the non-context clauses of $c_{i}$. For each $c_{i}\in CE$, its genuine pair $P^{gp}_{i}$ is obtained by the Cartesian product of the clauses in $CE$ and $IW_{i}$. Similarly, the fake pair $P^{fp}_{i}$ is obtained by the Cartesian product of the clauses in $CE$ and $OW_{i}$. 

Ideally, $CE$ contains all the emotion clauses, and $|w|$ is the farthest distance between the current emotion clause and the candidate cause clauses. The genuine pairs can contain all the emotion-cause pairs in the document, which means that the real emotion-cause pairs can be extracted directly from the genuine pairs. However, emotion prediction cannot find all the emotion clauses in the document. It compels us to equip EPO-ECPE with an ability to constantly adjust the results of emotion prediction during the training process by setting a synchronization to transfer the information between the emotion prediction and emotion-cause pair extraction. Specifically, this setting synchronizes the representation learned from the pair extraction to emotion prediction, so that emotion prediction can dynamically adjust the candidate emotion clauses based on the new clause representation. Once the performance of emotion prediction is further improved, the quality of $CE$ will be improved accordingly, thereby significantly reducing the difficulty of emotion-cause pair extraction. From this analysis, we propose the genuine pair and fake pair supervisions correspondingly update the clauses in the genuine pairs and fake pairs with different pair representations.

\subsection{Genuine Pair Supervision}
The genuine pair contains $K*(2*|w|+1)$ high-quality candidate emotion-cause pairs, where $K$ is the size of $CE$, $(2*|w|+1)$ is the context window size, and $K*(2*|w|+1)$ is less than $n*n$. We extract the emotion-cause pairs from the genuine pairs. The genuine pairs can be divided into three cases:
\begin{itemize}
\item Emotion-cause pairs
\item Mismatching pairs due to the wrong candidate emotion clause
\item Mismatching pairs due to the wrong cause clause
\end{itemize}
For these three cases, we adopt the weighted clause representations based on the relation of the emotion clause and cause clause to form the pair representation since a causal link exists between the cause and the emotion it triggers for a valid emotion-cause pair. 
Given a clause $c_{i}\in CE$, we take the clause $c_{j}\in IW_{i}$ as its candidate cause clause and then obtain their pair representation by the following equation:
\begin{equation}
{p}_{ij}^{gp} = {r}^{c}_{j}\odot \beta_{ij}\label{eq_p_gp}
\end{equation}
where $\beta_{ij} = \text{Softmax}({r}^{c}_{j}({r}^{e}_{i})^\mathsf{T})$, denoting the relevance of $c_{j}$ to $c_{i}$. ${r}^{c}_{j}$ is the context-specific representation of $c_{j}$, ${r}^{e}_{i}$ is the emotion-specific representation of $c_{i}$ and $\odot$ is the element multiplication. We use an MLP (parameterized by ${W}^{gp}$ and ${b}^{gp}$) with logistic function $\sigma(\cdot)$ to predict the probability of the pair $(c_{i},c_{j})$ being an emotion-cause pair (denoted as $\hat{y}_{ij}^{gp}$):
\begin{equation}
\hat{y}_{ij}^{gp} = \sigma({W}^{gp}{p}^{gp}_{ij} + {b}^{gp})
\end{equation}

The genuine pair supervision is trained by minimizing the cross entropy, which is given by:
\begin{small}
\begin{equation}
\mathcal{L}_{gp}= \sum_{i}^{|CE|}\sum_{j}^{|IW_{i}|}-(y_{ij}^{gp}log\hat{y}_{ij}^{gp} + (1-y_{ij}^{gp})log(1-\hat{y}_{ij}^{gp})) \label{eq_gp}
\end{equation}
\end{small}
where ${y}_{ij}^{gp}\in \{0,1\}$ is the ground-truth of pair $(c_{i},c_{j})$. Note that $y_{ij}^{gp}=1$ means that $(c_{i},c_{j})$ is an emotion-cause pair.
\subsection{Fake Pair Supervision} 
Obviously, the genuine pair supervision trains the pair representations of the genuine pairs, which indicates that not all clauses can participate in the training with the emotion-cause pair extraction, especially for the non-context clauses of the candidate emotion clause.Suppose that the genuine pair set still contains no emotion-cause pairs when we set the value of $|w|$ large enough to cover all cause clauses for each candidate emotion clause.
In this case, the emotion prediction stage gives incorrect candidate emotion clauses, and the real emotion clause drops in the fake pair set. Its clause representation can be updated with the help of fake pair supervision. As an auxiliary task, fake pair supervision is mainly used to assist the emotion prediction and genuine pair extraction in ensuring that each clause in the document can participate in the training of pair extraction. The concatenation of representations can preserve the original clause's features, which facilitates emotion prediction to receive feedback from the fake pair supervision. Therefore, we concatenate the representations of candidate emotion and non-context clauses as the pair representation:
\begin{equation}
{p}^{fp}_{ik} = [{r}^{e}_{i},{r}^{c}_{k}]
\end{equation}
where ${r}^{c}_{k}$ is the context-specific representation of clause $c_{k}\in OW_{i}$. After obtaining ${p}^{fp}_{ik}$, we use the prediction layer (parameterized by $\{{W}^{fp},{b}^{fp}\}$) to calculate the probability that pair $(c_{i},c_{k})$ is an emotion-cause pair:
\begin{equation}
\hat{y}_{ik}^{fp} = \sigma({W}^{fp}{p}^{fp}_{ik} + {b}^{fp})
\end{equation}
Similarly, the cross entropy of fake pair supervision is given as:
\begin{small}
\begin{equation}
\mathcal{L}_{fp}= \sum_{i}^{|CE|}\sum_{k}^{|OW_{i}|}-(y_{ik}^{fp}log\hat{y}_{ik}^{fp} + (1-y_{ik}^{fp})log(1-\hat{y}_{ik}^{fp}))\label{eq_fp}
\end{equation}
\end{small}
where ${y}_{ik}^{fp}\in \{0,1\}$ is the ground-truth of pair $(c_{i},c_{k})$.

\subsection{Training}
Considering that emotion prediction is critical to constructing genuine and fake pair sets, we first pre-train the emotion prediction module. We also pre-train the fake pair supervision since the feedback of fake pair supervision can be used to assist the emotion prediction. After the pre-training phase, the emotion prediction can achieve better performance before entering the training phase. Note that the more training steps in the pre-training phase are mainly to improve the emotion prediction performance instead of training emotion-cause pair extraction performance. Furthermore, the pre-training stage does not bring additional data or knowledge to EPO-ECPE performance. The total loss of emotion prediction and fake pair supervision is employed as the objective function $\mathcal{L}_{pre}$ of the pre-training phase:
\begin{equation}
\mathcal{L}_{pre} = \mathcal{L}_{e} +\mathcal{L}_{fp} \label{eq_pre}
\end{equation}
where $\mathcal{L}_{e}$ denotes the cross-entropy loss of emotion prediction.

In the training phase, we employ the sum of $\mathcal{L}_{e}$, $\mathcal{L}_{gp}$ and $\mathcal{L}_{fp}$ as the loss function $\mathcal{L}_{train}$ for the document $d$:
\begin{equation}
\mathcal{L}_{train} = \mathcal{L}_{e} + \mathcal{L}_{gp} + \mathcal{L}_{fp} \label{eq_tra}
\end{equation}
In the prediction phase, we modify the original lexicon-based extraction scheme \cite{DBLP:conf/acl/WeiZM20} and develop a new scheme to extract emotion-cause pairs from the genuine pairs. For each candidate emotion-cause pair $(c_{i}, c_{j})\in P^{gp}_{i}$, we set $(c_{i}, c_{j})$ to be an emotion-cause pair only if the following two conditions are both satisfied:
\begin{itemize}
\item[1.] Based on a sentiment lexicon, the candidate emotion clause $c_{i}$ is required to contain a sentiment word(s).  
\item[2.] $y^{gp}_{ij} > 0.5$ or the pair $(c_{i},c_{j})$ is the top pair with the highest score $y^{gp}_{ij}$.
\end{itemize}

\section{Experiment}
\begin{table}[t]
    \caption{Statistical information of the benchmark dataset, and we use the E-C pair to represent the emotion-cause pair.}
    \begin{center}
    \resizebox{\linewidth}{!}{\begin{tabular}{lccc}
    \hline
      &Item&Number&Percent(\%)\\
      \hline
       &Texts with one emotion-cause pair   &1746&89.77\\
       &Texts with two emotion-cause pairs   &77&9.10\\
       &Texts with more than two emotion-cause pairs   &22&1.13\\
       &The pairs with one emotion  &70&-\\
       &The pairs with more than one emotions   &129&-\\
       &Average of clause per text &14.77&-\\
       &Max of clause per text &73&-\\
       &Max number of emotion-cause pair per text &4&-\\
       \hline
    \end{tabular}}
    \end{center}
    \label{tab:si}
\end{table}

\subsection{Experimental Settings}
We use the benchmark dataset published by \cite{DBLP:conf/acl/XiaD19} to evaluate the proposed EPO-ECPE. This dataset is constructed based on emotion cause extraction corpus \cite{DBLP:conf/emnlp/GuiWXLZ16}, which contains 1,945 Chinese documents collected from SINA city news \footnote{ http://news.sina.com.cn/society/}. Specifically, 1,746 documents have one emotion-cause pair, 199 documents have two or more emotion-cause pairs, of which 70 documents contain only one emotion clause, and 129 documents contain multiple emotion clauses. Note that the emotion is certainly relevant to the causes. More details of the benchmark dataset can refer to Table \ref{tab:si}.
Following the previous work \cite{DBLP:conf/acl/WeiZM20}, we use 10-fold cross-validation to conduct experiments and conduct a one-sample t-test on the experimental results. We adopt the precision (P), recall (R), and F1 score (F1) as the metrics for evaluation and repeat the experiments 10 times to report the average result. Furthermore, we decompose the emotion-cause pairs to the emotion clause set and cause clause set to evaluate the performance of emotion clause extraction and cause clause extraction, respectively. 

\subsection{Implementation Details}
We adopt separately 200-dimension pre-trained Word2Vec \cite{DBLP:conf/nips/MikolovSCCD13}, and the pre-trained BERT encoder \cite{DBLP:conf/naacl/DevlinCLT19} to conduct experiments. For pre-trained Word2Vec, we use it to initialize the word embeddings. The pre-trained BERT encoder is initialized with BERT-Base, Chinese \footnote{https://github.com/google-research/bert}. The batch size and the dimension of clause representation are set to 32 and 200, respectively. We set the dropout rates of 0.1, 0.5, 0.1, and 0.1 for the embedding layer, word-level Bi-LSTM, clause-level Bi-LSTM, and prediction layer, respectively. In the pre-training phase, we set the learning rate and the number of pre-training rounds to 0.001 and 5. In the training phase, the learning rate and the number of training rounds are set to 0.001 and 50. The size of the emotion clause candidate set $CE$ and the value of $|w|$ are set to 3 (i.e., $K=3$) and 2, respectively. 
\begin{table*}[t]
   \caption{Experimental results of methods without using BERT as encoder on emotion-cause pair extraction, emotion extraction and cause extraction. The baseline results are reprinted from the corresponding publications.}
  \begin{center}
    \resizebox{\textwidth}{!}{\begin{tabular}{cccccccccc}\\
    \hline
  \multirow{2}*{}&
    \multicolumn{3}{c}{\bf Emotion-Cause Pair Extraction}&\multicolumn{3}{c}{\bf Emotion Extraction}&\multicolumn{3}{c}{\bf Cause Extraction}\cr\cline{2-10}
    &P&R&F1&P&R&F1&P&R&F1\cr
    \hline
    Inter-EC&0.6721&0.5705&0.6128&0.8364&0.8107&0.8230&0.7041&0.6083&0.6507\\
    E2EECPE&0.6478&0.6105&0.6280&0.8595&0.7915&0.8238&0.7062&0.6030&0.6503\\
    pairGCN&0.6999&0.5779&0.6321&0.8587&0.7208&0.7829&0.7283&0.5953&0.6541\\
    LML&0.6990&0.5960&0.6440&0.8810&0.7810&0.8260&-&-&-\\
    ECPE-2D&0.6960&0.6118&0.6496&0.8512&0.8220&0.8358&0.7272&0.6298&0.6738\\
    SLSN-U&0.6836&0.6291&0.6545&0.8406&0.7980&0.8181&0.6992&0.6588&0.6778\\
    RANKCP&0.6698&0.6546&0.6610&0.8703&0.8406&0.8548&0.6927&0.6743&0.6824\\
    ECPE-MLL&0.7090&0.6441&0.6740&0.8582&0.8429&0.8500&0.7248&0.6702&0.6950\\
    IE-CNN-CRF&0.7149&0.6279&0.6686&0.8614&0.7811&0.8188&0.7348&0.5841&0.6496\\
    UTOS&0.6911&0.6193&0.6524&0.8610&0.7925&0.8250&0.7189&0.6496&0.6802\\
    \hline
    EPO-ECPE(ours)&\textbf{0.7900}&0.6021&\textbf{0.6824}&\textbf{0.9780}&0.7848&\textbf{0.8702}&\textbf{0.7961}&0.6039&0.6848\\
    Standard Variance &0.0006&0.0009& 0.0007&-&-&-&-&-&-\\
    \hline
    \end{tabular}}
    \end{center}
    \label{tab:without_bert}
\end{table*}

\begin{table*}[t]
   \caption{Experimental results of methods using BERT as encoder on emotion-cause pair extraction, emotion extraction and cause extraction. The baseline results are reprinted from the corresponding publications.}
  \begin{center}
    \resizebox{\textwidth}{!}{\begin{tabular}{cccccccccc}\\
    \hline
  \multirow{2}*{}&
    \multicolumn{3}{c}{\bf Emotion-Cause Pair Extraction}&\multicolumn{3}{c}{\bf Emotion Extraction}&\multicolumn{3}{c}{\bf Cause Extraction}\cr\cline{2-10}
    &P&R&F1&P&R&F1&P&R&F1\cr
    \hline
    PairGCN&0.7692&0.6791&0.7202&0.8857&0.7958&0.8375&0.7907&0.6928&0.7375\\
    LMB&0.7110&0.6070&0.6550&0.8990&0.8000&0.8470&-&-&-\\
    ECPE-2D&0.7292&0.6544&0.6889&0.8627&0.9221&0.8910&0.7336&0.6934&0.7123 \\
    RANKCP&0.7119&0.7630&0.7360&0.9123&0.8999&0.9057&0.7461&0.7788&0.7615\\
    ECPE-MLL&0.7700&0.7235&0.7452&0.8608&0.9191&0.8886&0.7382&0.7912&0.7630\\
    Transition&0.7374&0.6307&0.6799&0.8716&0.8244&0.8474&0.7562&0.6471&0.6974\\
    Tagging&0.7243&0.6366&0.6776&0.8196&0.7329&0.7739&0.7490&0.6602&0.7018\\
    UTOS&0.7389&0.7062&0.7203&0.8815&0.8321&0.8559&0.7671&0.7320&0.7471\\
    Refinement&0.7377&0.6802&0.7078&0.8593&0.7993&0.8282&0.7614&0.7039&0.7315\\
    \hline
    EPO-ECPE(ours)&0.7621&0.7519&\textbf{0.7564}&\textbf{0.9787}&\textbf{0.9232}&\textbf{0.9500}&0.7711&0.7543&0.7620\\
    Standard Variance &0.0007&0.0004& 0.0005&-&-&-&-&-&-\\
    \hline
    \end{tabular}}
    \end{center}
    \label{tab:with_bert}
\end{table*}
\subsection{Baseline Methods}
To verify the effectiveness of the proposed EPO-ECPE, we compare it with 13 existing methods. We summarize these methods as follows:
\begin{itemize}
\item {\bf Inter-EC} is a pipeline model that extracts the emotion clauses and cause clauses separately and then extracts the pairs with causality based on the relative position between the emotion clause and cause clause \cite{DBLP:conf/acl/XiaD19}.
\item {\bf E2EECPE} takes the ECPE as a link prediction task and learns to make a link from the emotion clause to the cause clause, which means if there is a link between the emotion clause and cause clause, they can form an effective emotion-cause pair \cite{DBLP:journals/corr/abs-2002-10710}.
\item {\bf PairGCN} proposes a Pair Graph Convolutional Network model for the dependency among candidate pairs in the local neighborhood and considers the difference of dependency relations type in the propagation of contextual information \cite{DBLP:conf/coling/ChenHLWZ20}.
\item {\bf LAE-MANN} uses the multi-level attention mechanism based on LSTM or BERT encoders, which is denoted as \textbf{LML} and \textbf{LMB}, respectively \cite{DBLP:journals/ijon/TangJZ20}. 
\item {\bf ECPE-2D} uses a 2D-Transformer representation scheme to encode the inter-pair interactions under the window-constrained and cross-road 2D-Transformer to obtain the abundant information \cite{DBLP:conf/acl/DingXY20}. 
\item {\bf SLSN-U} detects and matches the emotion clause and cause clause simultaneously with the local search for extracting emotion-cause pairs \cite{DBLP:conf/coling/ChengJYYG20}.
\item {\bf RANKCP} \cite{DBLP:conf/acl/WeiZM20} proposes a unified framework to extract emotion-cause pairs from the ranking perspective by scoring the candidate pairs with consideration of dependency relations among clauses.
\item {\bf ECPE-MLL} is a multi-label joint framework, which contains the cause extraction corresponding to the specified emotion clause and the emotion extraction corresponding to the specified cause clause \cite{DBLP:conf/emnlp/DingXY20}.
\item {\bf Transition} views the ECPE as a procedure of parsing-like directed graph construction and generates the directed graph with labeled edges based on a sequence of actions \cite{DBLP:conf/acl/FanYDGYX20}.
\item {\bf Tagging} is a sequence labeling method, which takes the relative distance between the emotion clause and cause clause as the tag to ensure the emotion and the corresponding causes can be extracted simultaneously \cite{DBLP:conf/emnlp/YuanFBX20}.
\item {\bf IE-CNN-CRF} takes the emotional type as a part of the labeling tag to distinguish the cause clause under different emotional types to solve the extraction of multiple emotion-cause pairs in the document \cite{DBLP:conf/coling/ChenLW20}.
\item {\bf UTOS} views the emotion-cause pair extraction task as a unified sequence labeling problem. It divides the labeling tags into a content part for emotion/cause identification and a pairing part for clause pairing. Additionally, UTOS integrates the information of the target clause, global context, and former decoded label into an end-to-end unified sequence labeling framework \cite{DBLP:journals/taslp/ChengJYLG21}.
\item {\bf Refinement} proposes a multi-task sequence tagging framework, which encodes the distances between the emotion clause and cause clause into a novel tagging scheme to extract emotions with the associated causes simultaneously. It uses the prediction distribution of both auxiliary tasks as an inductive bias to refine the pair tagging distribution \cite{DBLP:journals/taslp/FanYGZX21}.
\end{itemize}
\subsection{Results and Analysis}
The existing methods' results can be divided into two categories: the results without BERT and the other with BERT. The results on the two categories are shown in Table \ref{tab:without_bert} and Table \ref{tab:with_bert}, respectively. The proposed EPO-ECPE achieves state-of-the-art performance compared with the existing methods, whether the BERT is used or not. 

First, we focus on the results of methods without using BERT (Table \ref{tab:without_bert}). Inter-EC yields low recall in the three tasks, especially for the emotion-cause pair extraction, which may be since Inter-EC is a two-step pipeline system. 
E2EECPE obtains significant improvement on the recall but a decrement in the precision for the emotion-cause pair extraction. SLSN-U and ECPE-MLL (the state-of-the-art method without using BERT) extract emotion-cause pairs by extracting cause clauses corresponding to the specified emotion clause and the emotion clauses corresponding to the specified cause clause in an end-to-end fashion, but ECPE-MLL performs better. It can be observed that EPO-ECPE achieves the best precisions in the three tasks and obtains 7.51\%, 9.7\%, and 6.13\% improvements, respectively. For the emotion-cause pair extraction, EPO-ECPE outperforms ECPE-MLL and achieves a new state-of-the-art performance. We randomly sampled 50\% data in each test set 20 times, used the trained model to predict the sample data results, and then obtained a sequence of F1 scores with a length of 20 to calculate the p-value. The improvements in accuracy are statistically significant with $p<0.01$. We notice that the recall drops happen in all three tasks compared with ECPE-MLL. The drops may be because the emotion-cause pairs are extracted only from the genuine pairs rather than from all possible pairs like the method ECPE-MLL. In other words, if the real emotion-cause pairs are not included in the genuine pairs, then the recall of emotion-cause pair extraction drops. The drop of recall of emotion-cause pair extraction definitely leads to the recall drops of the emotion extraction and cause extraction. 
Next, we focus on the results of methods with BERT (Table \ref{tab:with_bert}). The results show the F1 scores of EPO-ECPE are improved by 1.12\% and 6.14\% improvements in the emotion-cause pair extraction and emotion extraction, respectively, compared with the best-performing baseline ECPE-MLL using BERT as the encoder. We can find the results of EPO-ECPE are significantly better than that of the hierarchical encoder, especially in the recall rates of the three tasks. This result indicates the effectiveness of contextualized embeddings as external knowledge, and the pre-trained BERT is a suitable backbone encoder for clause representation learning. Furthermore, EPO-ECPE improves more significantly in the emotion clause extraction task, which may be thanks to the setting that EPO-ECPE is centered on emotion prediction and benefits from the synchronization mechanism between emotion prediction and pair extraction.
Meanwhile, the cause extraction and emotion-cause pair extraction also obtain improvements driven by the high performance of the emotion prediction. We also conduct a one-sample t-test on the F1 scores of emotion-cause pair extraction, and the improvements from 0.6740 to 0.6824 and 0.7452 to 0.7564 are statistically significant with $p<0.05$. We also calculate the standard variance of the metric for emotion-cause pair extraction and give the results in the last lines of Table \ref{tab:without_bert} and Table \ref{tab:with_bert}.
We speculate that the improvements are attributed to such mutual promotion among emotion prediction, genuine pair supervision, and fake pair supervision, which significantly exploits the potential of each part. These results illustrate that the proposed framework EPO-ECPE can accurately extract emotion-cause pairs without damaging the recall and further demonstrate the effectiveness of the synchronization settings presented in EPO-ECPE.

\begin{table*}[t]
    \caption{Examples of predicted emotion-cause pairs. The darker the color, the more likely the clause is to be used as a candidate emotional clause, while the lighter the color, the less likely the clause is to be used as a candidate emotional clause.}
    \begin{center}
    \begin{tabular}{llcc}
    \hline
    &Example&RANKCP Prediction&EPO-ECPE Prediction\\
    \hline
    &$[\cdots]$ $(c_{2})$\coderedmiddle{a man surnamed Chen suspected that his wife had an affair}, $(c_{3})$ after cutting his & \multirow{4}{*}{$(c_{2}, c_{3})$  \XSolidBrush} & \multirow{4}{*}{$(c_{5}, c_{4})$  \Checkmark}
    \\&wife more than 100 knives to death on the yesterday, $(c_{4}^{\S})$ he hanged himself near to the 
    \\&Taichung metropolitan park at night,$(c_{5}^{*})$\coderedhigh{leaving three children crying}, $[\cdots]$, ($c_{7}$)\coderedmiddle{Chen } \\&\coderedshallow{probably loves his wife too much}, $[\cdots]$ \\
    \hline
    &$(c_{1}^{*\S})$\coderedhigh{It's admirable to jump into the river to save people at the age of 66} $(c_{2})$ May the old & \multirow{4}{*}{$(c_{3}, c_{1})$  \XSolidBrush} & \multirow{4}{*}{$(c_{1}, c_{1})$ \Checkmark}
    \\& man be safe, $(c_{3})$\coderedshallow{Netizen Bian Guanyun praised 
    the righteous deeds of old man Chen}, 
    \\&$(c_{4})$ \coderedmiddle{Netizens worry about the safety of the elderly}, $[\cdots]$, $(c_{6})$hoping for a miracle, $[\cdots]$\\
    \hline
    &$[\cdots]$ $(c_{5})$\coderedshallow{but her words impressed the police}, $(c_{6})$ my father doesn't want me anymore,& \multirow{4}{*}{$(c_{8}, c_{8})$  \Checkmark} & \multirow{4}{*}{$(c_{8}, c_{8})$ \Checkmark}
    \\&$(c_{7})$ the first person to arrive at the police station was Yan
    Kuan, a mutual friend of \\&Wang Ling and her husband, $(c_{8}^{*\S})$\coderedhigh{he was shocked by Wang Ling's move to sink into }
    \\&\coderedhigh{the river with her daughter in her arms}, $[\cdots]$, $(c_{10})$\coderedmiddle{but I didn't expect her to be so upset}, $[\cdots]$\\
    \hline
    \end{tabular}
    \end{center}
    \label{tab:case_study}
\end{table*}

\subsection{Case Study}

We select some typical texts from the test set to conduct a case study for better analyzing how EPO-ECPE works in emotion-cause pair extraction. We also give the prediction results of RANKCP \cite{DBLP:conf/acl/WeiZM20}.
The results are shown in Table \ref{tab:case_study}. To visualize the results more clearly, we use the indications $c^{*}$ and $c^{\S}$ to represent the emotion clause and cause clause, respectively, and mark the candidate emotion clause with different color intensities to show the distribution of predicted candidate emotion clauses. 
In Example 1, RANKCP predicts clause $c_{2}$ as the emotion clause. One possible reason for this prediction result is that $c_{2}$ contains ``suspected'', which is usually viewed as an emotional word, which may confuse the prediction of RANKCP. Additionally, RANKCP is prone to predicting the clauses close to emotion as its cause since RANKCP incorporates the kernel-based relative position embedding in the ranking. This means that it is likely to lead to errors of cause extraction and emotion-cause pair extraction once the emotion clause is mispredicted. Compared with RANKCP, our model takes the clauses $c_{2}$, $c_{5}$ and $c_{7}$ as the candidate emotion clauses and then extracts the emotion-cause pairs from the genuine pairs constructed based on the candidate emotion clauses. It can be observed that the $c_{2}$ is also predicted as the emotion clause by EPO-ECPE, but its probability as the emotion clause is more minor than $c_{5}$. Similarly, this phenomenon also appears in Example 2. We can observe that $c_{3}$ is predicted as the emotion clause by RANKCP and the possible reason is that $c_{3}$ contains the word ``praised'', which is always regarded as an emotional word. Example 3 is a simple text with one emotion and one associated cause, and its emotion and cause occur in the same clause. In this case, it is easy to extract the emotion-cause pair, so both models can correctly extract $(c_{8},c_{8})$. The case study proved that emotion prediction plays a crucial role in solving ECPE task. Suppose the emotion clause can be correctly predicted. In that case, emotion-cause pair extraction (ECPE) can be regarded as emotion-cause extraction (ECE) to a certain extent, reducing extraction's difficulty.

\subsection{Error Analysis}
\begin{table*}[t]
    \caption{Examples of predicted emotion-cause pairs, where the first column depicts the content of the emotion-cause pair, the second column gives the ground-truth label, and the third column depicts the emotion-cause pairs identified by EPO-ECPE. The emotion clause is denoted with $c^{*}$ and the cause clause is denoted with $c^{\S}$.}
    \begin{center}
    \resizebox{\textwidth}{!}{\begin{tabular}{llcc}
    \hline
      &Emotion-cause Event&Ground Label&Prediction Label\\
       \hline
       &$[\cdots]$, ($c_{4}^{\S}$) now I have a baby too, ($c_{5}^{*}$) I'm so lucky, ($c_{6}^{*\S}$) and thank all those & \multirow{2}{*}{$(c_{5}, c_{4})$, $(c_{6}, c_{6})$}& \multirow{2}{*}{$(c_{5}, c_{4})$, $(c_{5}, c_{6})$}
       \\&who have helped me, ($c_{7}$) Cheng Xia is still lying in the hospital bed, $[\cdots]$\\
      \hline
       &$[\cdots]$, ($c_{2}^{\S}$) he has not married after thirty, ($c_{3}^{*}$) the family is very anxious, & \multirow{3}{*}{$(c_{3},c_{2})$, $(c_{10},c_{10})$}& \multirow{3}{*}{$(c_{10}, c_{10})$}
       \\&($c_{4}$) repeated blind dates and repeated rejections, $[\cdots]$, ($c_{10}^{*\S}$) Wang was very moved \\&because Chen didn't dislike his appearance, $[\cdots]$\\
      \hline
       &$[\cdots]$, ($c_{3}^{\S}$) there is a woman standing on the electric tower tottering, ($c_{4}^{*}$)The& \multirow{3}{*}{$(c_{4},c_{3})$}& \multirow{3}{*}{$(c_{9},c_{8})$}
       \\& villagers and passers-by were extremely anxious, $[\cdots]$, ($c_{8}$) a firefighter slowly climbed \\&to a place close to her 10 meters, ($c_{9}$) soothe her mood, $[\cdots]$\\
      \hline
      &($c_{1}$) According to Radio Hong Kong, $[\cdots]$, ($c_{14}$) judge described, ($c_{15}^{*}$)the case is a & \multirow{3}{*}{$(c_{15},c_{16})$}& \multirow{3}{*}{$(c_{16},c_{14})$,$(c_{16},c_{15})$}
      \\&sad tragedy, ($c_{16}^{\S}$) the stabbed defendant his wife with a sharp knife so hard that \\&his wife's sternum was broken and her main organs were seriously damaged, $[\cdots]$\\
      \hline
      &$[\cdots]$, ($c_{2}^{\S}$) It is found that the woman's appearance is very different from the photos & \multirow{2}{*}{$(c_{7},c_{2})$}& \multirow{2}{*}{$(c_{7},c_{8})$}
      \\&and her body shape is also very different, $[\cdots]$, ($c_{7}^{*}$) he is greatly disappointed,
      \\&$(c_{8})$ The man thought he had been deceived, $[\cdots]$\\
      \hline
      &$[\cdots]$, ($c_{2}$) the neurosurgeon explained that the patient had organ failure, & \multirow{4}{*}{$(c_{9},c_{3})$}& \multirow{4}{*}{$(c_{9},c_{8})$}
      \\&($c_{3}^{\S}$) the family members made a decision on organ donation, $[\cdots]$, ($c_{8}$) Meinong's relatives \\&and friends lament the early death of young life, ($c_{9}^{*}$) but I also admire the great love of \\&Da Xiong's family, $[\cdots]$\\
      \hline
    \end{tabular}}
    \end{center}
    \label{tab:error_case}
\end{table*}

To further understand the proposed model EPO-ECPE, we perform error analysis to observe what types of errors EPO-ECPE introduced or what kinds of emotion-cause pairs EPO-ECPE cannot be correctly extracted. The results are listed in Table \ref{tab:error_case}. We divide the extraction errors into emotion-cause pair extraction errors, emotion prediction errors, and cause prediction errors.

\subsubsection{Error Analysis of Emotion-Cause Pair Extraction} 
The error of emotion-cause pair extraction mainly refers to the wrong pairing relation between the emotion clause and the cause clause. This case usually appears in documents with multiple emotion-cause pairs. That is, although one is an emotion clause and the other is a cause clause, the two clauses do not match, so they cannot form an emotion-cause pair. For event 1, EPO-ECPE mismatches the emotion-cause pairing relation despite already detecting the emotion clause and cause clause, which indicates that our model is weak in encoding the causality of emotion cause. We speculate that this phenomenon may be because the pairwise representation encoder (i.e., Equation (\ref{eq_p_gp})) used in EPO-ECPE is relatively simple. We can observe from event 2 that EPO-ECPE can detect \{$c_{3}$,$c_{10}$\} as the candidate emotion clauses. Still, it is not always possible to recall all corresponding cause clauses for each candidate emotion clause, which damages the recall of emotion-cause pair extraction.

\subsubsection{Error Analysis of Emotion Prediction}
The error of emotion prediction mainly refers to the erroneous results caused by the wrong emotion clause prediction. From Table \ref{tab:error_case}, we find that $c_{4}$ is the emotion clause in event 3. However, our model detects $c_{9}$ as the emotion clause, which leads to the construction of an inaccurate emotion clause context window, and further leads to the extraction error of the emotion-cause pair. In event 4, EPO-ECPE incorrectly predicts $c_{16}$ as the emotion clause. It can be observed that $c_{16}$ is a long clause containing a lot of information, such as ``stabbed defendant'' and ``seriously damaged,'' which may disturb the emotion prediction module to determine the emotion clauses.

\subsubsection{Error Analysis of Cause Prediction}
The error of cause prediction mainly refers to the erroneous results caused by the wrong cause clause prediction. In events 5 and 6, our model can correctly predict the emotion clause but cannot recall the corresponding cause clause accurately. The main reason may be that our model treats the clauses close to the emotion clause as its context. Such a setting only allows the previous or following clauses of the candidate emotion clauses to be predicted as its candidate cause clauses. For event 5 and event 6, their cause clauses appear far from the emotion clause, so the actual cause clause will not be included in the context window of the emotion clause, resulting in prediction errors of the cause clause.

\subsection{Model Analysis}

\begin{table}[t]
    \caption{Comparison of effect of different supervision signals on emotion-cause pair extraction in pre-training and training phases.}
    \begin{center}
    \begin{tabular}{lllll}
    \hline
    &\multicolumn{1}{c}{}&\multicolumn{1}{c}{P}&\multicolumn{1}{c}{R}&\multicolumn{1}{c}{F1}\cr
      \hline
      \multirow{2}*{Pre-Training}
       &$\mathcal{L}_{fp}$      &0.7814&0.5839&0.6677\\
       &$\mathcal{L}_{e}$      &0.7083&0.5833&0.6395\\
       &$\mathcal{L}_{pre}$      &0.7288&0.5866&0.6497\\
      \hline
      \multirow{2}*{Training}
       &$\mathcal{L}_{fp}$      &0.7487&0.6033&0.6672\\
       &$\mathcal{L}_{e}$      &0.7359&0.5194&0.6054\\
      \hline
      \multirow{1}*{}
         &$Full$      &0.7900&0.6039&0.6824\\
        \hline
    \end{tabular}
    \end{center}
    \label{tab:sup_sig}
\end{table}
\subsubsection{Effect of Multiple Supervision}

The training of EPO-ECPE can be divided into pre-training and training phases. Each phase is mixed with multiple supervision signals, i.e., the signals $\mathcal{L}_{e}+\mathcal{L}_{fp}$ in the pre-training phase and $\mathcal{L}_{e}+\mathcal{L}_{gp}+\mathcal{L}_{fp}$ in the training phase. $\mathcal{L}_{e}+\mathcal{L}_{fp}$ aims to improve the performance of emotion prediction, and $\mathcal{L}_{e}+\mathcal{L}_{gp}+\mathcal{L}_{fp}$ mainly aims to improve the performance of emotion-cause pair extraction. We retrain EPO-ECPE after removing the specific supervision signal in the pre-training and training phases to verify the effects of different supervision signals. Experimental results are shown in Table \ref{tab:sup_sig}. In the pre-training phase, when $\mathcal{L}_{e}$ is removed, the F1 score drops to 0.6395. We speculate that the low-performance emotion prediction errors damage the quality of genuine pairs, leading to the drop of F1 in emotion-cause pair extraction. In addition, removing $\mathcal{L}_{fp}$ brings a degradation of nearly 1.5\% in the F1 score, which indicates that it should be necessary to introduce the signal of fake pair supervision in the pre-training phase. The fake pair and genuine pair supervisions constitute a closed representation learning structure, and removing any one of them causes the performance degradation. Furthermore, we remove the pre-training stage, and the F1 score of EPO-ECPE is reduced to 0.6497. We can observe that removing $\mathcal{L}_{pre}$ is less effective in the performance of emotion-cause pair extraction than removing $\mathcal{L}_{e}$. The reason for this phenomenon may be that the loss of pre-training $\mathcal{L}_{pre}$ is used to optimize the modules of emotion prediction and fake pair extraction simultaneously. The interaction of the two modules may lead to the improvement brought by the optimization of loss $\mathcal{L}_{pre}$ is not equal to the sum of improvements brought by optimizing $\mathcal{L}_{e}$ and $\mathcal{L}_{fp}$ separately. When $\mathcal{L}_{e}$ is removed for the training phase, the F1 scores decrease by 7.6\%. This drop illustrates that emotion prediction is crucial to the proposed model EPO-ECPE. We also eliminate $\mathcal{L}_{fp}$ in the training phase, which results in a 1.52\% drop in the F1 score. These results demonstrate the necessity of multiple supervision signals and the effectiveness of multiple supervision signals in the EPO-ECPE model, which helps learn the clause representations better, and further facilitates extracting emotion-cause pairs.


\begin{table}[t]
    \caption{Comparison with the existing results for documents with only one and more than one emotion-cause pair. The results of Inter-EC and RANKCP are implemented in \cite{DBLP:conf/acl/WeiZM20}.}
    \begin{center}
    \begin{tabular}{lccc}
    \hline
      &P&R&F\\
      \hline
       Single pair:\\
       Inter-EC     &0.6734&0.5939&0.6288\\
       RANKCP      &0.6625&0.6966&0.6780\\
       ECPE-MLL    &0.6870&0.6832&0.6851\\
       UTOS     &0.6765&0.6232&0.6480\\
       EPO-ECPE(ours)   &\textbf{0.7668}&0.6559&\textbf{0.7065}\\
       \hline
       Multiple pair:\\
       Inter-EC      &0.5912&0.3302&0.4206\\
       RANKCP      &0.7508&0.4390&0.5531\\
       ECPE-MLL    &0.7045&0.4776&0.5688\\
       UTOS     &0.5545&0.4676&0.5035\\
       EPO-ECPE(ours)      &\textbf{0.8396}&0.4768&\textbf{0.6019}\\
       \hline
    \end{tabular}
    \end{center}
    \label{tab:multi_pair}
\end{table}
\subsubsection{Comparison on Extracting Multiple Pairs}
We use the same experimental setting as \cite{DBLP:conf/acl/WeiZM20} to conduct experiments by dividing each fold's test set into single-pair documents and multiple-pair documents. Table \ref{tab:multi_pair} shows the experimental results on the two subsets. EPO-ECPE outperforms Inter-EC, RANKCP, and ECPE-MLL on both subsets. Compared with the best-performing model ECPE-MLL, the proposed EPO-ECPE obtains 3.3\% and 2.14 \% improvements in F1 measures for the single-pair extraction and multiple-pair extraction, respectively. It can be observed that compared with the competitive baselines, EPO-ECPE obtains a more noticeable improvement in multiple emotion-cause pair extraction. One possible reason for this result is that EPO-ECPE is a framework oriented toward emotion prediction. The multiple emotion-cause pairs can be simultaneously extracted from one document by matching the corresponding cause clauses for each possible emotion clause. Specifically, we construct the emotion-cause pair candidates for each candidate emotion clause to cover more possible emotion-cause pairs for a document so that we can easily extract all real emotion-cause pairs from those candidate pairs.
\begin{figure}[!t]
\centerline{\includegraphics[width=0.94\linewidth]{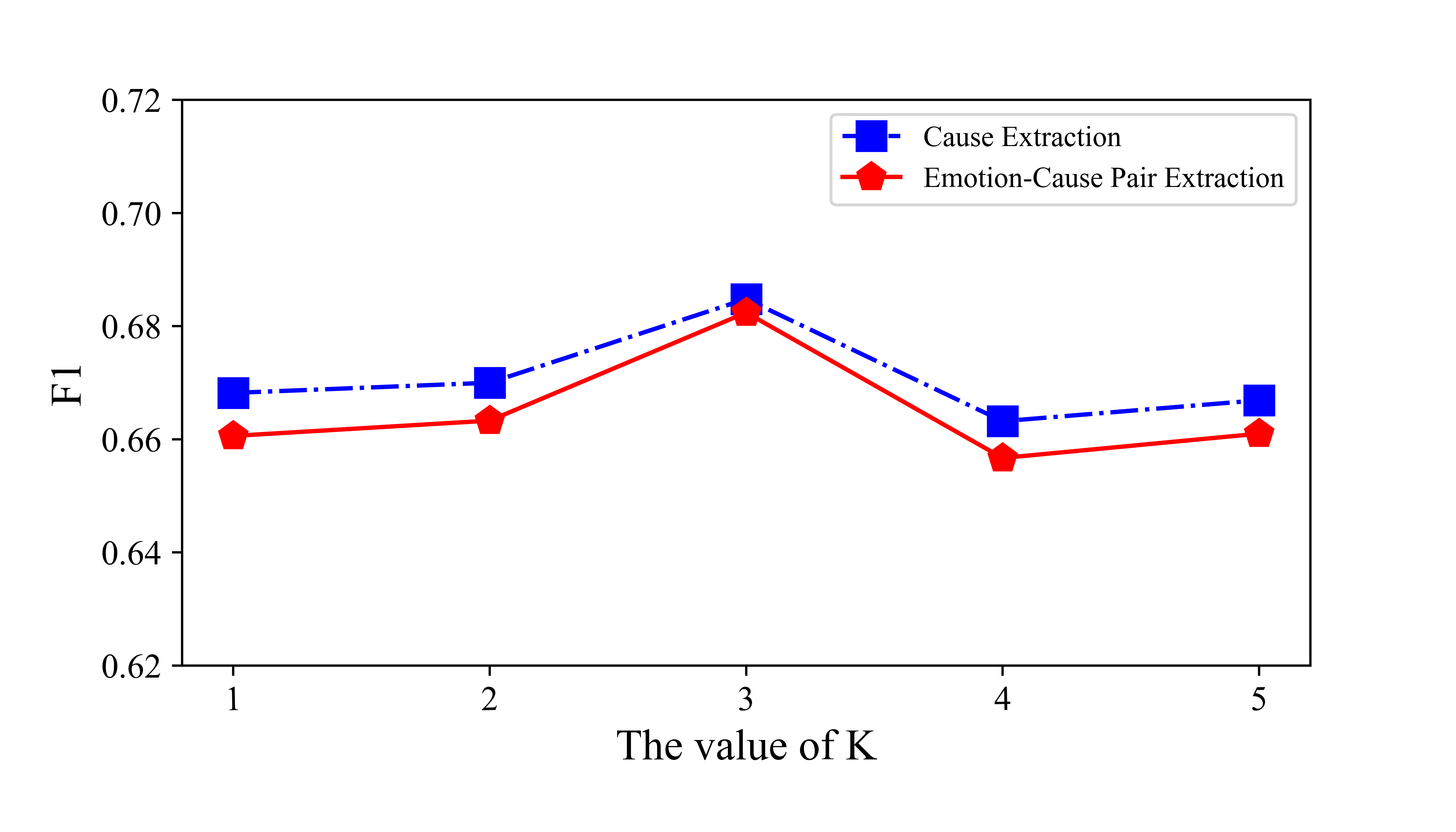}}
\caption{Results with various value of $K$.}
\label{fig:diff_k}
\end{figure}
\subsubsection{Effect of the Size of Emotion clause candidate set}
The quality of the emotion clause candidate set (i.e., $CE$) is bounded by the value of $K$. If $K$ is too large, it may contain some wrong clauses as emotion candidates. If $K$ is too small, it may not cover all valid emotion clauses in the document. Considering the two cases, we vary the value of $K$ (ranging from 1 to 5) to investigate the effects of $K$ on the cause extraction and emotion-cause pair extraction, as presented in Figure \ref{fig:diff_k}. From Figure \ref{fig:diff_k}, EPO-ECPE achieves the best performance on the two tasks when 3 clauses (i.e., $K$=3) are selected as the emotion clause candidates for each document, which is mainly determined by the characteristics of the dataset. If most documents in the dataset contain more emotional clauses, the value of K should be set larger. We can find from Table \ref{tab:si} that too large or too small $K$ will damage the performance of our model.
\begin{figure}[!t]
\centerline{\includegraphics[scale=0.48]{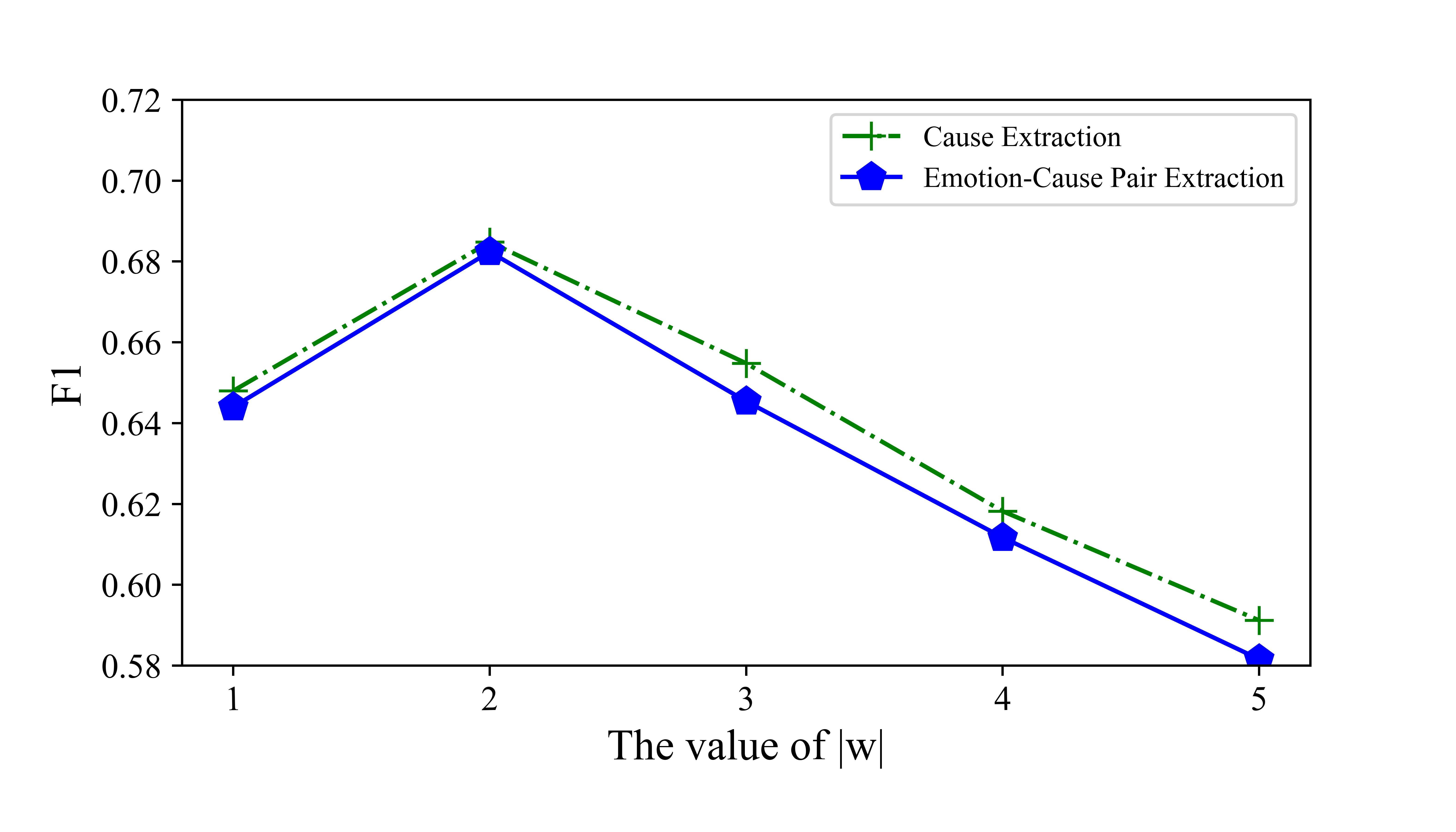}}
\caption{Results with various value of $|w|$.}
\label{fig:diff_w}
\end{figure}
\subsubsection{Effect of the Value of $|w|$}
$|w|$ is a hyperparameter to control which clause in the document can be considered the candidate cause clause for the given emotion clause. If $|w|$ is too large, it may introduce more wrong emotion-cause pairs into the genuine pairs, while $|w|$ is too small, it may not cover valid emotion-cause pairs. Both scenarios damage the performance of EPO-ECPE. Figure \ref{fig:diff_w} shows that EPO-ECPE achieves the best performance when $|w|=2$ (i.e., set 5 clauses as candidate cause clauses for each emotion clause). We find that 95.86\% of the relative distance between emotion and cause clause is 2 or less, so $|w|=2$ is reasonable for the benchmark dataset. 

\subsubsection{Generality of EPO-ECPE}
We can analyze the generality of EPO-ECPE from two aspects. First, the architecture of EPO-ECPE is not only suitable for ECPE tasks but also can be applied to handle other pairing problems, such as Pair-wise Aspect and Opinion Terms Extraction task (PAOTE). Specifically, the aspect extraction in PAOTE can be viewed as the emotion prediction to guide the pairing relation between the aspects and opinions. Then all aspect-opinion pairs can be divided into more likely and less likely to construct valid aspect-opinion pairs based on the aspect extraction results. The task decomposition promotes the joint learning of aspect and opinion extraction tasks. Second, EPO-ECPE’s idea of decomposing a difficult task into several slightly easier tasks can be applied to multi-task learning, especially when dependencies exist between multiple tasks. For such task, finding out which subtask is crucial to the main task can help each task make great use of the knowledge learned.

\section{Conclusions and Future Work}
In this paper, we propose an emotion prediction oriented framework (EPO-ECPE) to extract emotion-cause pairs, where the emotion prediction can guide the candidate pair construction and emotion-cause pair extraction. Specifically, we propose genuine and fake pair supervision to learn from the genuine and fake pairs, respectively. In this way, we build a synchronization mechanism to share their improvement in the training process. In the experimental section, EPO-ECPE significantly outperforms all existing methods and achieves state-of-the-art performance. Experimental results indicate that the proposed synchronization mechanism can help to better share the information between emotion prediction and emotion-cause pair extraction and further demonstrate the effectiveness of EPO-ECPE for solving ECPE task. In the future, we intend to dynamically integrate the contextual information based on whether the causality relation of the emotion-cause pair requires a specific context to avoid introducing noise.

\bibliography{manuscript} 
\bibliographystyle{manuscript}
\begin{IEEEbiography}[{\includegraphics[width=1.2in,height=1.4in,clip,keepaspectratio]{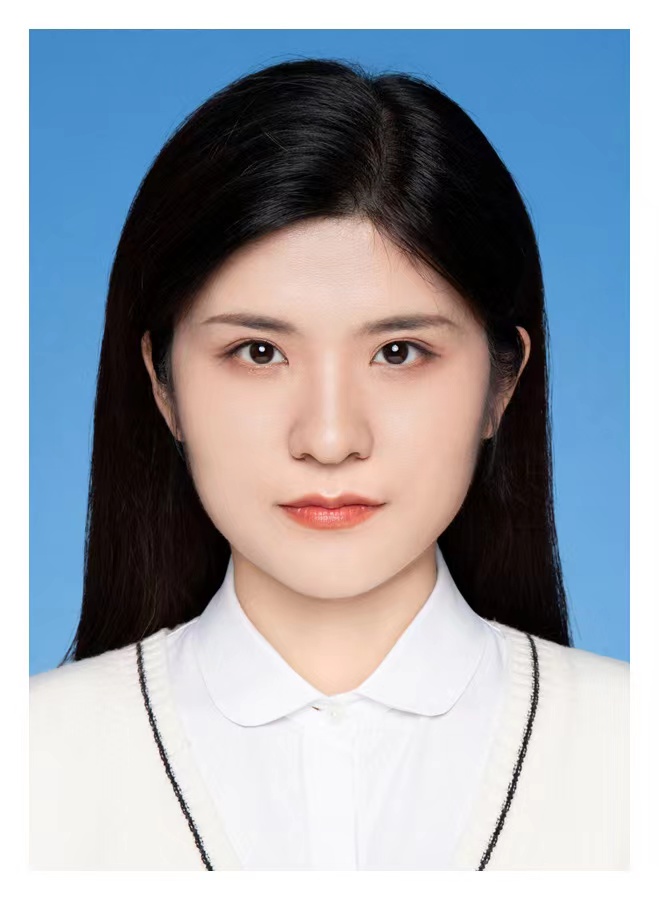}}]{Guimin Hu} received the B.S. degree in information management and information system from Liaocheng University, China, in 2015 and received the M.S. degree in computer science and technology from Northeastern University, China, in 2018. She is currently pursuing the Ph.D. degree with the Harbin Institute of Technology Shenzhen, China. Her current research interests lie at sentiment analysis, including emotion cause analysis, multi-modal sentiment analysis, and emotion recognition in conversation.
\end{IEEEbiography}
\begin{IEEEbiography}[{\includegraphics[width=1.1in,height=1.6in,clip,keepaspectratio]{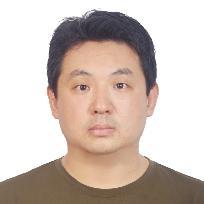}}]{Yi Zhao} received the Ph.D. degree in electronic engineering (nonlinear dynamics) from Hong Kong Polytechnic University, Hong Kong, in 2007. Since 2007, he has been with the Harbin Institute of Technology, Shenzhen,China, where he is currently a Professor. His research interests include applied dynamics, nonlinear time series analysis, and complex system modeling. His recent works have been on the application of mathematical methods to a diverse range of problems, including data science, biomathematics and the interpretability of the deep learning model.
\end{IEEEbiography}
\begin{IEEEbiography}[{\includegraphics[width=1in,height=1.25in,clip,keepaspectratio]{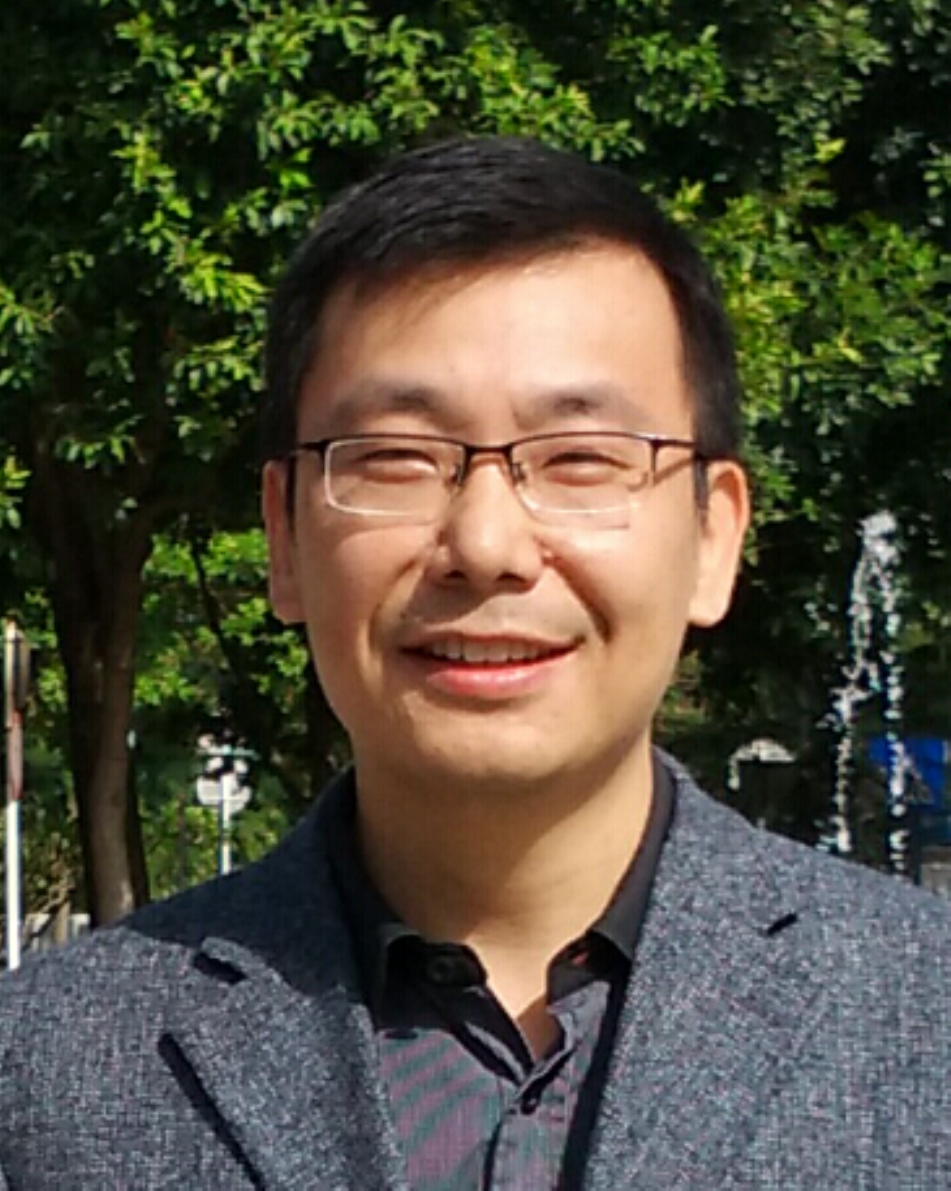}}]{Guangming Lu} received the B.S. degree in electrical engineering, the M.S. degree in control theory and control engineering, and the Ph.D. degree in computer science and engineering from the Harbin Institute of Technology, Harbin, China, in $1998$, $2000$, and $2005$, respectively. He is currently a Professor with Harbin Institute of Technology, Shenzhen, China. His current research interests include pattern recognition, image processing, and automated biometric technologies and applications.
\end{IEEEbiography}

\vfill

\end{document}